\documentclass{article}

\usepackage{graphicx}
\usepackage{pgfplots}
\pgfplotsset{compat=1.15}
\usepackage{mathrsfs}
\usetikzlibrary{arrows}
\usetikzlibrary{shapes.geometric}
\usetikzlibrary{positioning}
\usepackage{xcolor}
\usepackage{color}
\usepackage{amsfonts}
\usepackage{amsmath}
\usepackage{microtype}
\usepackage{graphicx}
\usepackage{subfigure}
\usepackage{booktabs} 

\usepackage{xcolor}

\usepackage{color}
  \definecolor{blueseven}{RGB}{0, 63, 92}
\definecolor{bluesix}{RGB}{55, 76, 128}
\definecolor{bluethree}{RGB}{122, 81, 149}
\definecolor{bluefour}{RGB}{188, 80, 144}
\definecolor{bluefive}{RGB}{239, 86, 117}
\definecolor{bluetwo}{RGB}{255, 118, 74}
\definecolor{blueone}{RGB}{255, 166, 0}
\definecolor{bluezero}{RGB}{100, 143, 255}
\definecolor{grayone}{RGB}{232, 236, 239}

\definecolor{bluenetone}{RGB}{122, 158, 252}

\definecolor{bluenettwo}{RGB}{210, 221, 241}

\usepackage{hyperref}

\usepackage[accepted]{icml2021}

\icmltitlerunning{Exact Spectral Norm Regularization for Neural Networks}

\newcommand\blfootnote[1]{%
  \begingroup
  \renewcommand\thefootnote{}\footnote{#1}%
  \addtocounter{footnote}{-1}%
  \endgroup
}

\begin{document}

\twocolumn[
\icmltitle{Exact Spectral Norm Regularization for Neural Networks}



\icmlsetsymbol{equal}{*}

\begin{icmlauthorlist}
\icmlauthor{Anton Johansson}{ch}
\icmlauthor{Niklas Engsner}{ch}
\icmlauthor{Claes Strannegård}{ch}
\icmlauthor{Petter Mostad}{ch}
\end{icmlauthorlist}

\icmlaffiliation{ch}{Chalmers University of Technology}

\icmlkeywords{Machine Learning, ICML}

\vskip 0.3in
]




\begin{abstract}



We pursue a line of research that seeks to regularize the spectral norm of the Jacobian of the input-output mapping for deep neural networks. While previous work rely on upper bounding techniques, we provide a scheme that targets the exact spectral norm. We showcase that our algorithm achieves an improved generalization performance compared to previous spectral regularization techniques while simultaneously maintaining a strong safeguard against natural and adversarial noise. Moreover, we further explore some previous reasoning concerning the strong adversarial protection that Jacobian regularization provides and show that it can be misleading.\blfootnote{$^1$Chalmers University of Technology. Correspondence to erikantonjohansson@gmail.com}
\end{abstract}

\section{Introduction}
\label{intro}

Ensuring that deep neural networks generalize can often be a question of applying the right regularization scheme. While long-established regularization schemes such as weight decay \cite{DBLP:conf/nips/KroghH91} can reduce the function complexity and prevent the network from overfitting, it can at times do so in a crude manner, reducing the complexity more than what is needed and inhibiting the overall performance of the network. Another important consideration for real-world generalizability that many regularization schemes fail to account for is robustness. Robustness will aid in ensuring that the model behaves as expected even when the input is perturbed, e.g., by natural or adversarial noise specifically crafted to fool a given model.
 With certain adversarial attack methods bridging the gap between theoretical concern and practical considerations by fooling commercial road signs detector with adversarial attacks \cite{DBLP:journals/corr/abs-1907-00374, DBLP:conf/pkdd/ChenCMC18}, robustness is becoming a progressively more important aspect of model deployment.

Previous work has demonstrated that regularizing the $l_p$-norms of the Jacobian of the network mapping can meet these two goals concurrently and different techniques have thus been developed to target these quantities \cite{DBLP:journals/tsp/SokolicGSR17}. Although obtaining the Jacobian is theoretically straightforward, it is computationally expensive and thus most schemes only seek to approximate a given norm. For example, the Frobenius norm has been approximated through sampling schemes and layer-wise approximations \cite{DBLP:journals/corr/abs-1908-02729, DBLP:journals/corr/GuR14} while the spectral norm has been targeted by upper-bounding the spectral norm of each weight matrix in the network \cite{ DBLP:journals/corr/YoshidaM17, DBLP:journals/tsp/SokolicGSR17}. 

In this work we extend on the schemes that target the spectral norm. While penalizing an upper-bound of the spectral norm does improve generalization and robustness, it is also crude in the sense that it does not directly target the quantity of interest and might thus inhibit the performance more than necessary. We instead provide an efficient algorithm that targets the \textit{exact} spectral norm of the Jacobian. Using this algorithm we demonstrate that targeting the exact spectral norm can yield an improved generalization performance while preserving a healthy defence against natural and adversarial perturbations.

\section{Background}

We follow \cite{DBLP:journals/corr/YoshidaM17} and represent an $L$-layer neural network $f: \mathbb{R}^{n_{in}} \rightarrow \mathbb{R}^{n_{out}}$ recursively as $x^l = f^l(G^l(x^{l-1}) + b^l),$ $l=1,2,...,L$ where $G^l$ is either a linear operator (e.g., convolution) or a piecewise linear operator (e.g., max-pool), $f^l$ the corresponding activation function, $b_l \in \mathbb{R}^{n_l}$ is the associated bias for layer $l$ and we set the input $x=x^0$. Denoting the collection of all parameters of the network as $\theta$, and making the dependence of the network on the parameters explicit as $f_{\theta}$, the full network function will be given as $f_{\theta}(x) = x^L$.


Momentarily restricting ourselves to the classification setting, the task that we are interested in is then the supervised learning problem of finding parameters $\theta$ such that $f_{\theta}$ can associate feature-values $x \in \mathbb{R}^{n_{in}}$ with one-hot encoded labels $y \in \mathbb{R}^{n_{out}}$ obtained from an unknown distribution $P$. This is achieved by collecting a training set $\mathcal{D}_t := \{(x_i,y_i)\}_{i=1}^N$ where $(x_i, y_i) \sim P$ and employing an appropriate loss function $l: \mathbb{R}^{n_{out}} \times \mathbb{R}^{n_{out}} \rightarrow \mathbb{R}$ which encourages $f_{\theta}$ to model a probability distribution for the possible labels for a given feature-value. Minimizing the full loss $l_{bare}(\theta, \mathcal{D}_t) := 1/|\mathcal{D}_t| \sum_{(x_i,y_i) \in \mathcal{D}_t} l(f_{\theta}(x_i), y_i)$ will thus align the distribution of $f_{\theta}(x_i)$ with that of the ground-truth label $y_i$. The minimization is done through some variant of stochastic gradient descent (SGD) where we split $\mathcal{D}_t$ into smaller disjoint random batches $\bigcup_i \mathcal{B}_i = \mathcal{D}_t$ and subsequently minimize $l_{bare}(\theta, \mathcal{D}_t)$ by reducing the partial loss $l_{\textrm{bare}}(\theta, \mathcal{B}_i)$ for every batch $\mathcal{B}_i$, whereupon the training set is split into new batches and the process repeated. We additionally utilize a validation set $\mathcal{D}_v := \{(x_i,y_i)\}_{i=1}^M$ with $(x_i,y_i) \sim P$ and $\mathcal{D}_t \cap \mathcal{D}_v = \emptyset$ to measure the performance of the model.

\subsection{Regularization}


Although the sole minimization of $l_{\textrm{bare}}(\theta, \mathcal{D}_t)$ can yield networks that perform adequately, the networks are often lacking in different regards such as generalization and robustness. While there exists a wide variety of methods that attempt to mitigate these deficiencies,
for example by controlling the magnitude of the weights as in weight decay, by utilizing knowledge distillation techniques \cite{DBLP:conf/wacv/AraniSZ21, DBLP:conf/sp/PapernotM0JS16} or by augmenting the training data with adversarially perturbed examples \cite{DBLP:conf/iclr/MadryMSTV18}, here we focus on the regularization techniques obtained by penalizing with some function $h: \mathbb{R} \rightarrow \mathbb{R}$ the norm of the Jacobian. This means that we seek to minimize
\begin{align}
l_{\textrm{jac}}(\theta, \mathcal{D}_t, \lambda) := &l_{\textrm{bare}}(\theta, \mathcal{D}_t) \label{eq:joint_loss_bare}\\
&+
\frac{\lambda}{|\mathcal{D}_t|} \sum_{(x_i, y_i) \in \mathcal{D}_t} h\bigg(\bigg|\bigg|\frac{df_{\theta}(x_i)}{dx}\bigg|\bigg|\bigg),\notag
\end{align}
where $\lambda$ is a hyper-parameter that controls the trade-off between the two terms and with typical choices for $h$ being either $h(x) = x$ or $h(x) = x^2$.

For most norms the regularized loss (\ref{eq:joint_loss_bare}) does not yield itself to any effective optimization schemes, requiring time-consuming operations to obtain the Jacobian for each $x_i$ in every batch $\mathcal{B}_i$. An exception to this is the Frobenius norm where one can obtain estimates either through a double-backpropagation scheme \cite{DBLP:journals/tnn/DruckerL92} or by using a more efficient sampling scheme \cite{DBLP:journals/corr/abs-1908-02729} where one samples $n_{proj}$ vectors $v^j$ from the $n_{out} - 1$ dimensional unit sphere $S^{n_{out} - 1}$ to approximate the squared Frobenius norm as 
\begin{align*}
    \bigg|\bigg|\frac{df_{\theta}(x)}{dx}\bigg|\bigg|^2_F &= n_{out}\mathbb{E}_{v \sim S^{n_{out} - 1}}\bigg[\bigg|\bigg|v \frac{df_{\theta}(x)}{dx}\bigg|\bigg|^2\bigg]\\
    &\approx \frac{n_{out}}{n_{proj}}\sum_{j=1}^{n_{proj}} \bigg[\frac{d (v^j\cdot x^L)}{d x}\bigg]^2,
\end{align*}
and thus minimizes the expression
\begin{align}
l_{\textrm{frob}}(\theta, \mathcal{D}_t, \lambda) &:= l_{\textrm{bare}}(\theta, \mathcal{D}_t) \label{eq:frobenius_loss} \\
&+ \frac{\lambda n_{out}}{|\mathcal{D}_t|n_{proj}} \sum_{(x_i, y_i) \in \mathcal{D}_t}\sum_{j=1}^{n_{proj}} \bigg[\frac{d (v^j\cdot x_i^L)}{d x}\bigg]^2.\notag
\end{align}

We on the other hand are interested in penalizing the spectral norm of the Jacobian at a point $x$, defined as
\begin{align}
    \bigg|\bigg|\frac{df_{\theta}(x)}{dx}\bigg|\bigg|_2 = \max_{\substack{ v\in\mathbb{R}^{n_{in}} \\ ||v||=1}} \bigg|\bigg|\frac{df_{\theta}(x)}{dx}v\bigg|\bigg|_2 = \sigma_{max}, \label{eq:spectral_jac}
\end{align}
where $\sigma_{max}$ denotes the largest singular value of $df_{\theta}(x)/dx$. A constraint on (\ref{eq:spectral_jac}) implies that we restrict the maximum rate at which $f_{\theta}$ can change as the input $x$ is perturbed, thus promoting robustness of our model. While the spectral norm does not immediately give itself to any viable method, \cite{DBLP:journals/corr/YoshidaM17} managed to develop an efficient scheme by restricting themselves to the setting where all activation functions are piecewise linear. Networks with piecewise linear activation functions are themselves piecewise linear functions and the input space can thus be decomposed into a partition $\mathcal{R}$ where for each $R \in \mathcal{R}$ there exists $W_R \in \mathbb{R}^{n_{in} \times n_{out}}$, $b_R \in \mathbb{R}^{n_{out}}$ such that $f_{\theta}(x) = W_Rx + b_R,~ \forall x \in R$ \cite{DBLP:conf/nips/HaninR19}. For these piecewise linear networks, the Jacobian $d f_{\theta}/dx$ is constant in each region $R \in \mathcal{R}$ and given by $W_R$. Calculating the spectral norm of the Jacobian at some input $x$ is thus reduced to calculating the spectral norm of $W_R$ associated with $R \ni x$. 

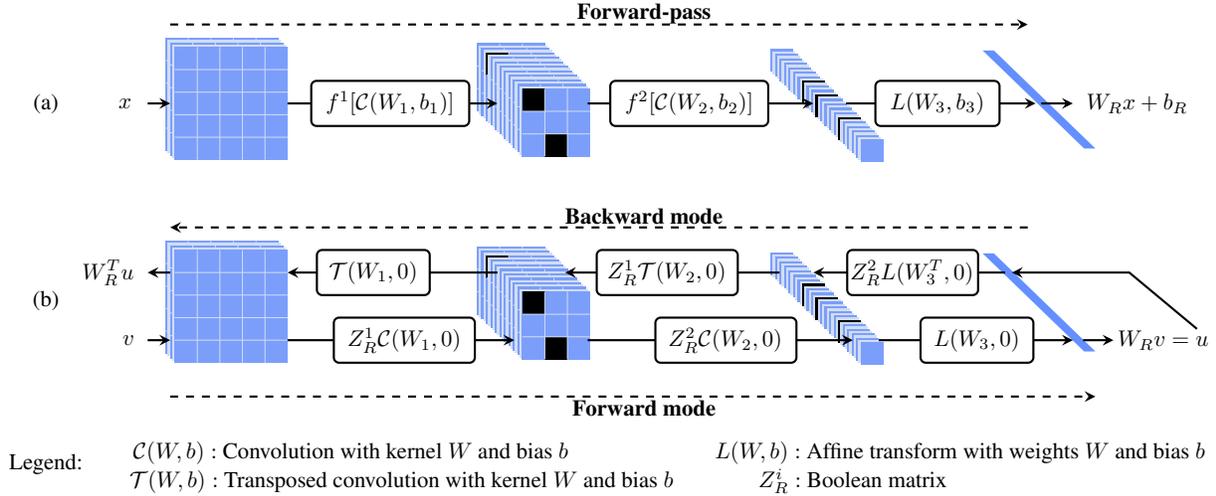
\begin{figure*}[h]
\centering

\begin{tikzpicture} [transform shape, scale = 1.5]

\tikzstyle{arrow} = [thick,->,>=stealth]

\newcommand\shiftAll{1.8}

\newcommand\imgWidth{0.2}

\newcounter{num}

\setcounter{num}{0}
\newcommand\numC{5}
\newcommand\chanW{1.4}
\newcommand\lcx{-1.7}
\newcommand\lcy{-0.6}
\newcommand\opac{0.99}
\newcommand\shift{0.02}

\newcommand\gridW{0.2}

\newcommand\scaleText{0.57}

\setcounter{num}{0}
\renewcommand\numC{5}
\renewcommand\chanW{1.0}
\renewcommand\lcx{0.14}
\renewcommand\lcy{-0.4 + \shiftAll}
\renewcommand\opac{0.99}
\renewcommand\shift{0.02}

\foreach \x in {1,...,\numC}
{

    \ifodd \value{num}
    {
    \fill[bluenettwo, opacity=\opac] (\lcx + \x*\shift, \lcy - \x*\shift) -- (\lcx + \chanW + \x*\shift, \lcy - \x*\shift) -- (\lcx + \chanW +  \x*\shift, \lcy + \chanW - \x*\shift) -- (\lcx + \x*\shift, \lcy + \chanW - \x*\shift) -- cycle;
    }\else 
    {
   \fill[bluenetone, opacity=\opac] (\lcx + \x*\shift, \lcy - \x*\shift) -- (\lcx + \chanW + \x*\shift, \lcy - \x*\shift) -- (\lcx + \chanW +  \x*\shift, \lcy + \chanW - \x*\shift) -- (\lcx + \x*\shift, \lcy + \chanW - \x*\shift) -- cycle;
    }\fi
    
    

    \foreach \g in {1,...,4}{
        \draw[bluenettwo, line width=0.01pt] (\lcx + \x*\shift + \g*\gridW, \lcy - \x*\shift) -- (\lcx + \x*\shift + \g*\gridW, \lcy + \chanW - \x*\shift);
        
        \draw[bluenettwo, line width=0.01pt] (\lcx + \x*\shift, \lcy - \x*\shift + \g*\gridW) -- (\lcx + \x*\shift + \chanW, \lcy + \g*\gridW - \x*\shift);
    }
    
    \addtocounter{num}{1}
    
}


\setcounter{num}{0}
\renewcommand\numC{21}
\renewcommand\chanW{0.6}
\renewcommand\lcx{2.9}
\renewcommand\lcy{-0.05 + \shiftAll}
\renewcommand\opac{0.99}
\renewcommand\shift{0.02}

\foreach \x in {1,...,\numC}
{

    \ifodd \value{num}
    {
    \fill[bluenettwo, opacity=\opac] (\lcx + \x*\shift, \lcy - \x*\shift) -- (\lcx + \chanW + \x*\shift, \lcy - \x*\shift) -- (\lcx + \chanW +  \x*\shift, \lcy + \chanW - \x*\shift) -- (\lcx + \x*\shift, \lcy + \chanW - \x*\shift) -- cycle;
    }\else 
    {
   \fill[bluenetone, opacity=\opac] (\lcx + \x*\shift, \lcy - \x*\shift) -- (\lcx + \chanW + \x*\shift, \lcy - \x*\shift) -- (\lcx + \chanW +  \x*\shift, \lcy + \chanW - \x*\shift) -- (\lcx + \x*\shift, \lcy + \chanW - \x*\shift) -- cycle;
    }\fi
    
    \fill[black] (\lcx + \numC*\shift + 0*\gridW, \lcy - \numC*\shift + 2*\gridW) -- (\lcx + \numC*\shift + 1*\gridW, \lcy - \numC*\shift + 2*\gridW) -- (\lcx + \numC*\shift + 1*\gridW, \lcy - \numC*\shift + 3*\gridW) -- (\lcx + \numC*\shift + 0*\gridW, \lcy - \numC*\shift + 3*\gridW) -- cycle;
    
    \fill[black] (\lcx + \numC*\shift + 1*\gridW, \lcy - \numC*\shift + 0*\gridW) -- (\lcx + \numC*\shift + 2*\gridW, \lcy - \numC*\shift + 0*\gridW) -- (\lcx + \numC*\shift + 2*\gridW, \lcy - \numC*\shift + 1*\gridW) -- (\lcx + \numC*\shift + 1*\gridW, \lcy - \numC*\shift + 1*\gridW) -- cycle;
    
    \ifnum \x=5
    \fill[black] (\lcx + 5*\shift + 0*\gridW, \lcy - 5*\shift + 2*\gridW) -- (\lcx + 5*\shift + 1*\gridW, \lcy - 5*\shift + 2*\gridW) -- (\lcx + 5*\shift + 1*\gridW, \lcy - 5*\shift + 3*\gridW) -- (\lcx + 5*\shift + 0*\gridW, \lcy - 5*\shift + 3*\gridW) -- cycle;
    \fi
    
    
    \foreach \g in {1,...,2}{
        \draw[bluenettwo, line width=0.01pt] (\lcx + \x*\shift + \g*\gridW, \lcy - \x*\shift) -- (\lcx + \x*\shift + \g*\gridW, \lcy + \chanW - \x*\shift);
        
        \draw[bluenettwo, line width=0.01pt] (\lcx + \x*\shift, \lcy - \x*\shift + \g*\gridW) -- (\lcx + \x*\shift + \chanW, \lcy + \g*\gridW - \x*\shift);
    }
    
    \addtocounter{num}{1}
    
}


\setcounter{num}{0}
\renewcommand\numC{41}
\renewcommand\chanW{0.2}
\renewcommand\lcx{5.5}
\renewcommand\lcy{0.3 + \shiftAll}
\renewcommand\opac{0.99}
\renewcommand\shift{0.02}

\foreach \x in {1,...,\numC}
{

    \ifodd \value{num}
    {
    \fill[bluenettwo, opacity=\opac] (\lcx + \x*\shift, \lcy - \x*\shift) -- (\lcx + \chanW + \x*\shift, \lcy - \x*\shift) -- (\lcx + \chanW +  \x*\shift, \lcy + \chanW - \x*\shift) -- (\lcx + \x*\shift, \lcy + \chanW - \x*\shift) -- cycle;
    }\else 
    {
   \fill[bluenetone, opacity=\opac] (\lcx + \x*\shift, \lcy - \x*\shift) -- (\lcx + \chanW + \x*\shift, \lcy - \x*\shift) -- (\lcx + \chanW +  \x*\shift, \lcy + \chanW - \x*\shift) -- (\lcx + \x*\shift, \lcy + \chanW - \x*\shift) -- cycle;
    }\fi
    
    \ifnum \x=15
    \fill[black] (\lcx + 15*\shift + 0*\gridW, \lcy - 15*\shift + 0*\gridW) -- (\lcx + 15*\shift + 1*\gridW, \lcy - 15*\shift + 0*\gridW) -- (\lcx + 15*\shift + 1*\gridW, \lcy - 15*\shift + 1*\gridW) -- (\lcx + 15*\shift + 0*\gridW, \lcy - 15*\shift + 1*\gridW) -- cycle;
    \fi
    
    \ifnum \x=21
    \fill[black] (\lcx + 21*\shift + 0*\gridW, \lcy - 21*\shift + 0*\gridW) -- (\lcx + 21*\shift + 1*\gridW, \lcy - 21*\shift + 0*\gridW) -- (\lcx + 21*\shift + 1*\gridW, \lcy - 21*\shift + 1*\gridW) -- (\lcx + 21*\shift + 0*\gridW, \lcy - 21*\shift + 1*\gridW) -- cycle;
    \fi
    
    \ifnum \x=31
    \fill[black] (\lcx + 31*\shift + 0*\gridW, \lcy - 31*\shift + 0*\gridW) -- (\lcx + 31*\shift + 1*\gridW, \lcy - 31*\shift + 0*\gridW) -- (\lcx + 31*\shift + 1*\gridW, \lcy - 31*\shift + 1*\gridW) -- (\lcx + 31*\shift + 0*\gridW, \lcy - 31*\shift + 1*\gridW) -- cycle;
    \fi

    \addtocounter{num}{1}
    
}

\fill[bluezero] (8.3,-0.4 + \shiftAll) -- (7.4, 0.47 + \shiftAll) -- (7.5, 0.47 + \shiftAll) -- (8.4,-0.4 + \shiftAll) -- cycle;

\draw[arrow] (1.24,0 + \shiftAll) -- (3.1,0 + \shiftAll);

\filldraw[draw=black, fill=white, rounded corners=2, thick] (1.45,-0.2 + \shiftAll) rectangle ++(1.4,0.4);

\node[scale = \scaleText] at (2.15, 0 + \shiftAll) {$f^1[\mathcal{C}(W_1,b_1)]$};



\draw[arrow] (3.9,0 + \shiftAll) -- (5.9,0 + \shiftAll);

\filldraw[draw=black, fill=white, rounded corners=2, thick] (4.1,-0.2 + \shiftAll) rectangle ++(1.4,0.4);

\node[scale = \scaleText] at (4.8, 0 + \shiftAll) {$f^2[\mathcal{C}(W_2,b_2)]$};


\draw[arrow] (6.19,0 + \shiftAll) -- (7.87,0 + \shiftAll);

\filldraw[draw=black, fill=white, rounded corners=2, thick] (6.45,-0.2 + \shiftAll) rectangle ++(1.1,0.4);

\node[scale = \scaleText] at (7.0, 0 + \shiftAll) {$L(W_3,b_3)$};

\draw[arrow] (7.92,0 + \shiftAll) -- (8.2,0 + \shiftAll);
\node[scale = \scaleText] at (8.77, 0 + \shiftAll) {$W_Rx + b_R$};

\draw[arrow] (0,\shiftAll) -- (0.2,\shiftAll);
\node[scale = \scaleText] at (-0.2, 0 + \shiftAll) {$x$};



\setcounter{num}{0}
\renewcommand\numC{5}
\renewcommand\chanW{1.4}
\renewcommand\lcx{-1.7}
\renewcommand\lcy{-0.6}
\renewcommand\opac{0.99}
\renewcommand\shift{0.02}

\renewcommand\gridW{0.2}


    
    
        

    

\setcounter{num}{0}
\renewcommand\numC{5}
\renewcommand\chanW{1.0}
\renewcommand\lcx{0.14}
\renewcommand\lcy{-0.4}
\renewcommand\opac{0.99}
\renewcommand\shift{0.02}

\foreach \x in {1,...,\numC}
{

    \ifodd \value{num}
    {
    \fill[bluenettwo, opacity=\opac] (\lcx + \x*\shift, \lcy - \x*\shift) -- (\lcx + \chanW + \x*\shift, \lcy - \x*\shift) -- (\lcx + \chanW +  \x*\shift, \lcy + \chanW - \x*\shift) -- (\lcx + \x*\shift, \lcy + \chanW - \x*\shift) -- cycle;
    }\else 
    {
   \fill[bluenetone, opacity=\opac] (\lcx + \x*\shift, \lcy - \x*\shift) -- (\lcx + \chanW + \x*\shift, \lcy - \x*\shift) -- (\lcx + \chanW +  \x*\shift, \lcy + \chanW - \x*\shift) -- (\lcx + \x*\shift, \lcy + \chanW - \x*\shift) -- cycle;
    }\fi
    
    

    \foreach \g in {1,...,4}{
        \draw[bluenettwo, line width=0.01pt] (\lcx + \x*\shift + \g*\gridW, \lcy - \x*\shift) -- (\lcx + \x*\shift + \g*\gridW, \lcy + \chanW - \x*\shift);
        
        \draw[bluenettwo, line width=0.01pt] (\lcx + \x*\shift, \lcy - \x*\shift + \g*\gridW) -- (\lcx + \x*\shift + \chanW, \lcy + \g*\gridW - \x*\shift);
    }
    
    \addtocounter{num}{1}
    
}


\setcounter{num}{0}
\renewcommand\numC{21}
\renewcommand\chanW{0.6}
\renewcommand\lcx{2.9}
\renewcommand\lcy{-0.05}
\renewcommand\opac{0.99}
\renewcommand\shift{0.02}

\foreach \x in {1,...,\numC}
{

    \ifodd \value{num}
    {
    \fill[bluenettwo, opacity=\opac] (\lcx + \x*\shift, \lcy - \x*\shift) -- (\lcx + \chanW + \x*\shift, \lcy - \x*\shift) -- (\lcx + \chanW +  \x*\shift, \lcy + \chanW - \x*\shift) -- (\lcx + \x*\shift, \lcy + \chanW - \x*\shift) -- cycle;
    }\else 
    {
   \fill[bluenetone, opacity=\opac] (\lcx + \x*\shift, \lcy - \x*\shift) -- (\lcx + \chanW + \x*\shift, \lcy - \x*\shift) -- (\lcx + \chanW +  \x*\shift, \lcy + \chanW - \x*\shift) -- (\lcx + \x*\shift, \lcy + \chanW - \x*\shift) -- cycle;
    }\fi
    
    \fill[black] (\lcx + \numC*\shift + 0*\gridW, \lcy - \numC*\shift + 2*\gridW) -- (\lcx + \numC*\shift + 1*\gridW, \lcy - \numC*\shift + 2*\gridW) -- (\lcx + \numC*\shift + 1*\gridW, \lcy - \numC*\shift + 3*\gridW) -- (\lcx + \numC*\shift + 0*\gridW, \lcy - \numC*\shift + 3*\gridW) -- cycle;
    
    \fill[black] (\lcx + \numC*\shift + 1*\gridW, \lcy - \numC*\shift + 0*\gridW) -- (\lcx + \numC*\shift + 2*\gridW, \lcy - \numC*\shift + 0*\gridW) -- (\lcx + \numC*\shift + 2*\gridW, \lcy - \numC*\shift + 1*\gridW) -- (\lcx + \numC*\shift + 1*\gridW, \lcy - \numC*\shift + 1*\gridW) -- cycle;
    
    \ifnum \x=5
    \fill[black] (\lcx + 5*\shift + 0*\gridW, \lcy - 5*\shift + 2*\gridW) -- (\lcx + 5*\shift + 1*\gridW, \lcy - 5*\shift + 2*\gridW) -- (\lcx + 5*\shift + 1*\gridW, \lcy - 5*\shift + 3*\gridW) -- (\lcx + 5*\shift + 0*\gridW, \lcy - 5*\shift + 3*\gridW) -- cycle;
    \fi
    
    
    \foreach \g in {1,...,2}{
        \draw[bluenettwo, line width=0.01pt] (\lcx + \x*\shift + \g*\gridW, \lcy - \x*\shift) -- (\lcx + \x*\shift + \g*\gridW, \lcy + \chanW - \x*\shift);
        
        \draw[bluenettwo, line width=0.01pt] (\lcx + \x*\shift, \lcy - \x*\shift + \g*\gridW) -- (\lcx + \x*\shift + \chanW, \lcy + \g*\gridW - \x*\shift);
    }
    
    \addtocounter{num}{1}
    
}


\setcounter{num}{0}
\renewcommand\numC{41}
\renewcommand\chanW{0.2}
\renewcommand\lcx{5.5}
\renewcommand\lcy{0.3}
\renewcommand\opac{0.99}
\renewcommand\shift{0.02}

\foreach \x in {1,...,\numC}
{

    \ifodd \value{num}
    {
    \fill[bluenettwo, opacity=\opac] (\lcx + \x*\shift, \lcy - \x*\shift) -- (\lcx + \chanW + \x*\shift, \lcy - \x*\shift) -- (\lcx + \chanW +  \x*\shift, \lcy + \chanW - \x*\shift) -- (\lcx + \x*\shift, \lcy + \chanW - \x*\shift) -- cycle;
    }\else 
    {
   \fill[bluenetone, opacity=\opac] (\lcx + \x*\shift, \lcy - \x*\shift) -- (\lcx + \chanW + \x*\shift, \lcy - \x*\shift) -- (\lcx + \chanW +  \x*\shift, \lcy + \chanW - \x*\shift) -- (\lcx + \x*\shift, \lcy + \chanW - \x*\shift) -- cycle;
    }\fi
    
    \ifnum \x=15
    \fill[black] (\lcx + 15*\shift + 0*\gridW, \lcy - 15*\shift + 0*\gridW) -- (\lcx + 15*\shift + 1*\gridW, \lcy - 15*\shift + 0*\gridW) -- (\lcx + 15*\shift + 1*\gridW, \lcy - 15*\shift + 1*\gridW) -- (\lcx + 15*\shift + 0*\gridW, \lcy - 15*\shift + 1*\gridW) -- cycle;
    \fi
    
    \ifnum \x=21
    \fill[black] (\lcx + 21*\shift + 0*\gridW, \lcy - 21*\shift + 0*\gridW) -- (\lcx + 21*\shift + 1*\gridW, \lcy - 21*\shift + 0*\gridW) -- (\lcx + 21*\shift + 1*\gridW, \lcy - 21*\shift + 1*\gridW) -- (\lcx + 21*\shift + 0*\gridW, \lcy - 21*\shift + 1*\gridW) -- cycle;
    \fi
    
    \ifnum \x=31
    \fill[black] (\lcx + 31*\shift + 0*\gridW, \lcy - 31*\shift + 0*\gridW) -- (\lcx + 31*\shift + 1*\gridW, \lcy - 31*\shift + 0*\gridW) -- (\lcx + 31*\shift + 1*\gridW, \lcy - 31*\shift + 1*\gridW) -- (\lcx + 31*\shift + 0*\gridW, \lcy - 31*\shift + 1*\gridW) -- cycle;
    \fi

    \addtocounter{num}{1}
    
}

\fill[bluezero] (8.3,-0.4) -- (7.4, 0.47) -- (7.5, 0.47) -- (8.4,-0.4) -- cycle;






\draw[arrow] (1.24,-0.3) -- (3.25,-0.3);

\filldraw[draw=black, fill=white, rounded corners=2, thick] (1.65,-0.5) rectangle ++(1.25,0.4);

\node[scale = \scaleText] at (2.275, -0.3) {$Z_R^1\mathcal{C}(W_1,0)$};

\draw[arrow] (3.1,0.3) -- (1.24,0.3) ;

\filldraw[draw=black, fill=white, rounded corners=2, thick] (1.5,0.1) rectangle ++(1.0,0.4);

\node[scale = \scaleText] at (2.0, 0.3) {$\mathcal{T}(W_1,0)$};




\draw[arrow] (3.9,-0.3) -- (6.25,-0.3);

\filldraw[draw=black, fill=white, rounded corners=2, thick] (4.5,-0.5) rectangle ++(1.25,0.4);

\node[scale = \scaleText] at (5.12, -0.3) {$Z_R^2\mathcal{C}(W_2,0)$};

\draw[arrow]  (5.6,0.3) -- (3.7,0.3);

\filldraw[draw=black, fill=white, rounded corners=2, thick] (3.98,0.1) rectangle ++(1.25,0.4);

\node[scale = \scaleText] at (4.6, 0.3) {$Z_R^1\mathcal{T}(W_2,0)$};


\draw[arrow] (6.48,-0.3) -- (8.2,-0.3);

\filldraw[draw=black, fill=white, rounded corners=2, thick] (6.85,-0.5) rectangle ++(1.0,0.4);

\node[scale = \scaleText] at (7.35, -0.3) {$L(W_3,0)$};

\draw[arrow] (7.6,0.3) -- (5.9,0.3) ;

\filldraw[draw=black, fill=white, rounded corners=2, thick] (6.2,0.1) rectangle ++(1.16,0.4);

\node[scale = \scaleText] at (6.775, 0.3) {$Z_R^2L(W_3^T,0)$};

\draw[arrow] (8.29,-0.3) -- (8.56,-0.3);
\node[scale = \scaleText] at (9.0, -0.3) {$W_Rv = u$};

\draw[arrow] (9.3,-0.2) -- (8.7, 0.3) -- (7.66, 0.3);

\draw[arrow] (0,-0.3) -- (0.2,-0.3);
\node[scale = \scaleText] at (-0.17, -0.3) {$v$};

\draw[arrow] (0.2,0.3) -- (0,0.3);
\node[scale = \scaleText] at (-0.35, 0.3) {$W_R^Tu$};

\node[scale = \scaleText] at (4.4, 2.6) {\textbf{Forward-pass}};
\draw[arrow, dashed] (0.2,2.5) -- (7.8,2.5);

\node[scale = \scaleText] at (4.4, -0.9) {\textbf{Forward mode}};
\draw[arrow, dashed] (0.2,-0.8) -- (8.4,-0.8);

\node[scale = \scaleText] at (4.4, 0.8) {\textbf{Backward mode}};
\draw[arrow, dashed] (7.8,0.7) -- (0.2,0.7);

\node[scale = \scaleText] at (1.8, -1.3) {$\mathcal{C}(W,b):$ Convolution with kernel $W$ and bias $b$};

\node[scale = \scaleText] at (2.25, -1.55) {$\mathcal{T}(W,b):$ Transposed convolution with kernel $W$ and bias $b$};

\node[scale = \scaleText] at (6.25, -1.55) {$Z_R^i:$ Boolean matrix};

\node[scale = \scaleText] at (7.2, -1.3) {$L(W,b):$ Affine transform with weights $W$ and bias $b$};

\node[scale = \scaleText] at (-0.9, 1.8) {(a)};

\node[scale = \scaleText] at (-0.9, 0.05) {(b)};

\node[scale = \scaleText] at (-0.9, -1.4) {Legend:};

\end{tikzpicture}

\caption{The difference between a regular forward-pass and the forward and backward modes for a two hidden layer network. (a) A regular forward-pass of $x$ through the network. Each box showcases the operation that maps the input between the layers. The black squares indicate the neurons mapped to zero by the ReLU activation functions $f^i$. (b) The forward and backward modes used to estimate $||W_R||_2$. An input $v$ is sent through the network to yield $u$ whereupon $u$ is sent backwards through the network. Note how each operation is now bias-free with the same weights as during the forward-pass.
The activation functions are replaced by multiplication with the Boolean matrices designed to keep the activation pattern fixed, see equation (\ref{eq:wr_forw_back_interp1}) - (\ref{eq:wr_forw_back_interp2}). The backward mode is achieved through transposed convolutions and linear transformations.}
    \label{fig:full_illu_bf}
\end{figure*}

Although the regularization scheme is valid for all piecewise linear activation functions, it is easiest to present for networks with only ReLU \cite{DBLP:conf/icml/NairH10} activation functions and we thus momentarily restrict ourselves to this setting. By restricting ourselves to these networks and by using the fact that all linear and piecewise linear operators $G^l$ can locally be represented as a matrix $W^l$, one can obtain the identity \begin{align}
    W_R = W^LZ_R^{L-1}W^{L-1}\cdots W^2Z_R^1W^1
    \label{eq:wr_identity}
\end{align}
 where $Z_R^i$ is a diagonal boolean matrix indicating which neurons in layer $i$ that have an output $>0$ when passing $x\in R$ through the network. Using this identity, an upper bound for $||W_R||_2$ can be obtained as $||W_R||_2\leq \prod_l ||W^l||_2$ and subsequently \cite{DBLP:journals/corr/YoshidaM17} regularize the spectral norm by bounding the spectral norm of each weight matrix. They thus minimize the expression
\begin{align}
    l_{\textrm{specUB}}(\theta, \mathcal{D}_t, \lambda) := l_{\textrm{bare}}(\theta, \mathcal{D}_t) + \lambda\sum_{l=1}^L ||W^l||^2_2,
    \label{eq:loss_yoshida}
\end{align}
and additionally suggest to further effectivize the scheme by using power iteration on the matrices $W^l$ as $v \sim S^{n_l - 1}$, $u \leftarrow W^lv$, $v \leftarrow (W^l)^Tu$ to approximate the spectral norm as $||W^l||_2 \approx ||u||_2/||v||_2$. While this scheme will penalize the spectral norm of the Jacobian, it only does so through an upper bound, thus potentially inhibiting the performance of the network more than necessary.

\section{Method}
Here we introduce our method which penalizes the spectral norm of the Jacobian directly. Our scheme relies on power iteration as previous methods but targets $||W_R||_2$ directly.
We will follow prior research and momentarily restrict ourselves to piecewise linear networks without skip-connections since this provides a scheme that is easy to present and implement,
but keep in mind that the ensuing methodology is valid for networks with skip-connection as well. Additionally, the scheme can be extended efficiently to networks utilizing any non-linear transformation at the cost of a slightly more involved implementation scheme. We detail this extension scheme in Section \ref{sec:extens}.

\subsection{Exact spectral norm regularization}

To perform power iteration on $W_R$ we need a way to efficiently perform the steps $v \sim S^{n_{in} - 1}$, $u \leftarrow W_Rv$, $v \leftarrow W_R^Tu$ to subsequently approximate the norm as $||W_R||_2 \approx ||u||_2/||v||_2$. Given that the main obstacle for an efficient scheme is the construction of $W_R$, our scheme circumvents the construction by directly focusing on the matrix-vector products $W_Rv$ and $W_R^Tu$. Returning to the identity (\ref{eq:wr_identity}), we can see that, given the constituent weight matrices $W^l$ and boolean matrices $Z_R^l$, one can obtain the desired matrix-vector products as\begin{align}
    W_Rv &= W^LZ_R^{L-1}W^{L-1}\cdots W^2Z_R^1W^1v\label{eq:wr_forw_back_interp1},\\
    W_R^Tu &= (W^1)^TZ_R^1(W^2)^T\cdots (W^{L-1})^TZ_R^{L-1}(W^L)^Tu.
    \label{eq:wr_forw_back_interp2}
\end{align} While the matrices $Z_R^l$ can easily be obtained by recording which neurons that have an output $>0$ when passing $x\in R$ through the network, the construction of the matrices $W^l$ is inefficient for most network layers except for the very simplest ones, making the direct application of (\ref{eq:wr_forw_back_interp1}) - (\ref{eq:wr_forw_back_interp2}) impractical. 

While the direct application is impractical, we can obtain a practical scheme by interpreting equations (\ref{eq:wr_forw_back_interp1}) - (\ref{eq:wr_forw_back_interp2}) in a particular manner. Equation (\ref{eq:wr_forw_back_interp1}) is nothing other than the forward-pass of $v$ through the network \textit{with all bias vectors set to 0} and \textit{the activation functions replaced with multiplication with boolean matrices $Z_R^l$}, hereby referred to as the \textit{forward mode} of the network. Similarly, equation (\ref{eq:wr_forw_back_interp2}) is the output obtained by passing $u$ \textit{backwards} through the network, meaning that we start at the final layer and transform $u$ layer by layer with analogous modifications to the bias vectors and activation functions as in the forward mode until we reach the input layer. 
We will hereby refer to this reverse pass as the \textit{backward mode}\footnote{Note that the backward mode can be obtained by a standard backward pass to evaluate $d(x^L \cdot u)/dx$. We refer to it here as backward mode to highlight the symmetry with the forward mode which does not have a standard equivalent counterpart.} of the network. This interpretation circumvents the formation of the matrices $W^l$ and instead relies on forward and backward operators $F^l$ and $(F^L)^T$ that make use of the linear and piecewise linear operators $G^l$ and their corresponding transposed version $(G^l)^T$ that implicitly define $W^l$ and $(W^l)^T$, e.g., through convolution and transposed convolution operators. While for many layers we have that the layer transformations $G^l$ and the resulting forward operators $F^l$ coincide, meaning $F^l = G^l$, there do exist some exceptions to this rule where a little bit of extra care is needed to ensure that the
forward and backward modes correctly map to $W_Rv$ and $W_R^Tu$ respectively, e.g., max-pooling layers where the max-indices of the forward-pass has to be utilized. The reader is referred to the Appendix to see the conversion between the operators $G^l, F^l$ and $(F^l)^T$ for some commonly used layers.

Thus we can target the exact spectral norm of $W_R$ by performing power iteration with $v \sim S^{n_{in} - 1}$ and obtain the matrix-vector products $W_Rv$ and $(W_R)^Tu$ through the forward and backward mode respectively, thereupon estimating the spectral norm as $||W_R||_2 \approx ||u||_2/||v||_2$\footnote{It is possible to perform power iteration multiple times to get a better estimate but we found that performing it once gave sufficiently accurate estimates.}. For a visualization of the difference between a regular forward-pass, the forward and backward mode of the network and the involved operators, see Figure \ref{fig:full_illu_bf} where all of this is visualized for a simple three layer convolutional network. The network only
utilizes ReLU activation functions so that $f^1=f^2=$ReLU and $G^1,G^2$ are given by convolutional layers while $G^3$ is a linear layer. 

 Making the association between $R$ and an input $x, x\in R$, explicit as $R_x$, we can formulate our exact spectral loss as
\begin{align}
    l_{\textrm{spec}}(\theta, \mathcal{D}_t, \lambda) &:= l_{\textrm{bare}}(\theta, \mathcal{D}_t) \label{eq:loss_spectral}\\
    &+ \frac{\lambda}{|\mathcal{D}_t|}\sum_{(x_i,y_i) \in \mathcal{D}_t} ||W_{R_{x_i}}||_2. \notag
\end{align}
Further, converting the matrix multiplication with the Boolean matrices $Z_R^i$ to component-wise Hadamard products $\odot$ with vectors $z^i$, we can formulate the entire scheme on a batch level which can be seen in Algorithm 1.

\begin{algorithm}[tb]
   \caption{Spectral norm regularization}
   \label{alg:spec_reg}
\begin{algorithmic}
   \STATE {\bfseries Input:} Mini-batch $\mathcal{B}_i$ of feature-value pairs $(x,y)$, weight factor $\lambda$, number of power iterations $N$
   \STATE {\bfseries Output:} Approximate gradient $\nabla_{\theta }l_{spec}(\theta, \mathcal{B}_i, \lambda)$
   \STATE $x^0 = x$ \COMMENT{Forward-pass start}
   \FOR{$l=1$ {\bfseries to} $L$} 
   \STATE $x^l = f^l(G^l(x^{l-1}) + b^l)$ 
   \IF{$l < L$}
   \STATE $z^l = \mathbb{I}\{x^l > 0\}$
   \ENDIF
   \ENDFOR
   \STATE $v \sim \mathcal{N}(0,I)$ \COMMENT{$v$ is of shape $(|\mathcal{B}_i|, n_{in})$}
   \STATE $v = v/||v||_2$ \COMMENT{Normalize rows}
   \FOR{$n=1$ {\bfseries to} $N$} 
   \STATE \COMMENT{Forward-mode start}
   \FOR{$l=1$ {\bfseries to} $L$} 
   \STATE $v = F^l(v)$ 
   \IF{$l<L$}
   \STATE $v = v \odot z^l$
   \ENDIF
   \ENDFOR
   \STATE $u = v$
   \STATE $u = u/||u||_2$ \COMMENT{Normalize rows}
   \STATE \COMMENT{Backward-mode start}
   \FOR{$l=L$ {\bfseries to} $1$} 
   \STATE $u = (F^l)^T(u)$
   \IF{$l>1$}
   \STATE $u = u \odot z^{l-1}$
   \ENDIF
   \ENDFOR
   \ENDFOR
   \STATE $\sum_{(x_i, y_i) \in \mathcal{B}_i}||W_{R_{x_i}}||_2 = \textrm{sum}(||u||_2/||v||_2)$
   \STATE $R_{spec}(\theta) = \sum_{(x_i, y_i) \in \mathcal{B}_i}||W_{R_{x_i}}||_2$
   \STATE $\nabla_{\theta} l_{spec}(\theta, \mathcal{B}_i, \lambda) = \nabla_{\theta} l_{\textrm{bare}}(\theta, \mathcal{B}_i) + \nabla_{\theta} \frac{\lambda}{|\mathcal{B}_i|}R_{spec}(\theta)$
\end{algorithmic}
\end{algorithm}

\subsection{Extension to non-piecewise linear transforms}
\label{sec:extens}
While the scheme detailed in Algorithm 1 is capable of regularizing the spectral norm of the Jacobian, it is easiest to implement and most efficient in the piecewise linear setting where all layer-wise transformations are given by piecewise linear functions. Although this is a restriction, many well performing networks rely solely on non-linearities given by piecewise linear activation functions with the addition of batch-normalization layers, see for example VGG \cite{DBLP:journals/corr/SimonyanZ14a} and ResNet \cite{DBLP:conf/cvpr/HeZRS16} among others. Creating an easily implementable regularization scheme for this well-performing setting thus only requires us to additionally ensure the validity of the scheme when using batch-normalization.

Batch-normalization poses two issues which complicates the extension of the regularization scheme.
\begin{enumerate}
    \item Division by the variance of the input makes batch-normalization a non-piecewise linear transformation during training.
    \item Since the mean and variance are calculated per batch, batch-normalization induces a relation between input $x_j$ and output $f_{\theta}(x_i)$ where $x_i, x_j \in \mathcal{B}, i \neq j$.
    This induced relation adds multiple components $df_{\theta}(x_i)/dx_j$
    to the Jacobian which represents how an input $x_j$ affects an output $f_{\theta}(x_i)$. We believe these components are not relevant in practice and effort should thus not be spent controlling them.
\end{enumerate}
Since both of these issues are only present during training, we circumvent them by penalizing the spectral norm of the Jacobian obtained by
momentarily engaging a pseudo-inference mode where we set the running mean and variance of the batch-normalization layers to be fixed and given by the variance and mean obtained from the batch.

\begin{figure}[h]
    \centering
    \includegraphics[width=0.47\textwidth]{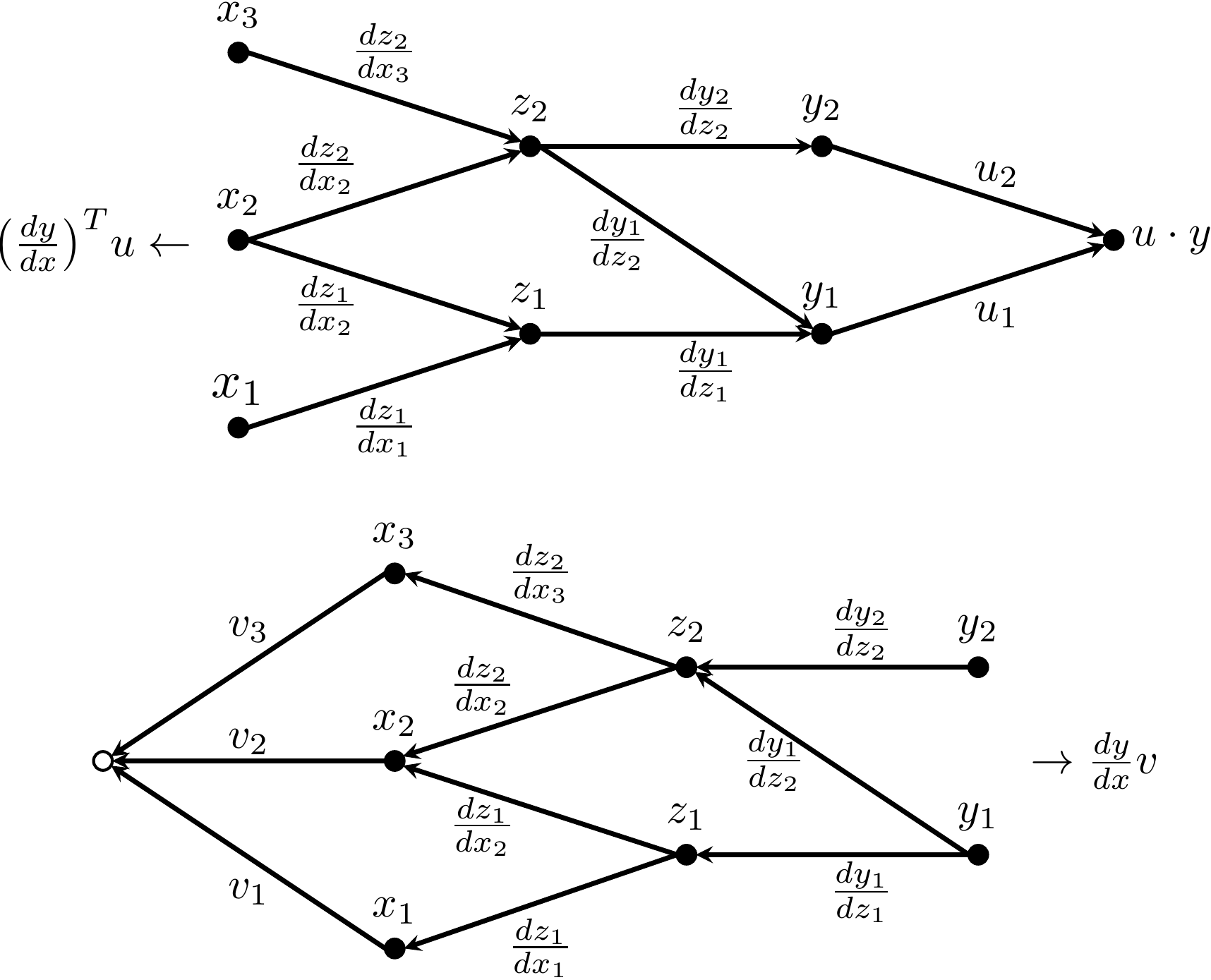}
    \caption{Illustration of the extension scheme. (Top) The computational graph associated with the forward-pass. Each node can perform any non-linear transformation of the associated input. To obtain the $i$:th component of $(dy/dx)^Tu$ we sum the product of the edge elements along every path from the right-most node to $x_i$. (Bottom) The modified computational graph to obtain $(dy/dx)v$. The direction of all edges are flipped, we remove the right-most node and we add a fictitious node to the computational graph (marked as a empty circle) with connecting edge elements being given by components of $v$. All other relevant edge elements are preserved from the top graph. The $i$:th component of $(dy/dx)v$ can then be obtained by starting at the fictitious node and summing the product of the edge elements along every path to $y_i$
    }
    \label{fig:ext}
\end{figure}

Additionally, Algorithm 1 can be efficiently extended to networks employing non-piecewise linear transformations as well at the cost of a more complicated implementation scheme. While not explicitly stated, Algorithm 1 can be used for non-piecewise linear transformations, replacing the Boolean matrices $Z^l$ with matrices given by $df^l/dx^{l-1}$. However, the calculations and storage of these matrices is likely to be cumbersome and memory intensive for most naive implementations and networks and we thus recommend that one instead utilizes the internal computational graph present in most deep learning libraries.
Calculating equation (\ref{eq:wr_forw_back_interp2}) is equivalent to calculating $(df/dx)^Tu$ and can thus be obtained by simply applying back-propagation to $d(x^L \cdot u)/dx$ which is a valid scheme for all networks, not only piecewise linear ones. Similarly we can obtain the matrix-vector product $(df/dx)v$ by utilizing the same computational graph used to obtain $d(x^L \cdot u)/dx$, but reverse the direction of all relevant constituent edges and adding a fictitious node to represent the inner product with $v$, see Figure \ref{fig:ext} for a demonstration of this fact for a simple computational graph. We relegate the proof of this to the Appendix.

We choose not to
focus on this possible implementation further though since the piecewise linear setting already encompasses a large amount of models and we believe that most will find the scheme in Algorithm 1 more straightforward to implement than delving deep into the mechanics of computational graphs. Further adding on to this fact is that the internals of the computational graphs of popular deep learning frameworks (such as PyTorch \cite{DBLP:conf/nips/PaszkeGMLBCKLGA19} and TensorFlow \cite{tensorflow2015-whitepaper}) are written in C++ and having to perform modifications of the graph would thus potentially impede the Python-based workflow which many practitioners operate with. However, if one wishes to utilize spectral regularization for networks that employ non-piecewise linear activation functions, for example sigmoids which can be of relevance for attention mechanisms \cite{DBLP:conf/nips/VaswaniSPUJGKP17}, then the extension scheme provides a well-principled and efficient approach that one can follow. In that case one would replace the Forward-mode and Backward-mode in Algorithm 1 with the computational graph manipulation techniques to obtain $(df/dx)v$ and $(df/dx)^Tu$ respectively.

\section{Experiments}

In this section we evaluate how targeting the exact spectral norm, hereby referred to as the Spectral method, compares to other regularization methods, namely the Frobenius method of (\ref{eq:frobenius_loss}), the Spectral-Bound method of (\ref{eq:loss_yoshida}) and weight decay \cite{DBLP:conf/nips/KroghH91} (also referred to as L2-regularization). We compare the generalization performance across different data sets and investigate the robustness of the obtained networks. 
\subsection{Generalization}
The considered data sets where the generalization performance is measured are KMNIST \cite{DBLP:journals/corr/abs-1812-01718}, FashionMNIST (which at times we will abbreviate as FMNIST) \cite{DBLP:journals/corr/abs-1708-07747} and CIFAR10 \cite{Krizhevsky09learningmultiple}. The generalization performance is measured by measuring the accuracy on the corresponding validation set $\mathcal{D}_v$ for each data set. A variant of the LeNet architecture \cite{lenet} is used for the KMNIST and FMNIST dataset while the VGG16 \cite{DBLP:journals/corr/SimonyanZ14a} architecture is used for CIFAR10. All
three data sets are preprocessed so that they have channelwise mean of 0 and a standard deviation of 1. We perform a grid-search to find the optimal hyperparameters for each network and regularization scheme, see the Appendix for more details regarding the training setup. Each experiment is repeated five times and the model that has the lowest mean loss over all hyperparameters over these five runs is chosen as the representative of a given method. The results of this experiment can be seen in Table \ref{table:gen_res}.
\begin{table}[h]
\caption{Mean test accuracy $\pm$ one standard deviation for the different regularization methods on three data sets computed over 5 runs. Bold indicates best mean accuracy. The method names have been shortened to make the table more compact.}
\label{table:gen_res}
\vskip 0.15in
\begin{center}
\begin{small}
\begin{sc}
\begin{tabular}{lcccr}
\toprule
Method & CIFAR10 & KMNIST & FMNIST \\
\midrule
Spec & 90.20 $\pm$ 0.61  & \textbf{96.61} $\pm$ \textbf{0.14} & \textbf{91.10} $\pm$ \textbf{0.06}\\
Frob & \textbf{90.21} $\pm$ \textbf{0.69} & 96.39 $\pm$ 0.08 & 91.00 $\pm$ 0.16\\
Spec-B & 89.37 $\pm$ 0.70 & 95.72 $\pm$ 0.21 & 90.66 $\pm$ 0.26\\
L2 & 89.94 $\pm$ 0.76 & 95.58 $\pm$ 0.05 & 90.64 $\pm$ 0.30 \\
 None & 88.59 $\pm$ 0.67 &  94.36 $\pm$ 0.26 & 90.35 $\pm$ 0.28\\
\bottomrule
\end{tabular}
\end{sc}
\end{small}
\end{center}
\vskip -0.1in
\end{table}

From these results we can see that penalizing the exact spectral norm on KMNIST and FMNIST does result in models with higher accuracies than those obtained from models with other regularization schemes, and for CIFAR10 it results in the second best model when considering the mean accuracy.
For KMNIST and FMNIST we can additionally see that the Spectral method is significantly better than the Spectral-Bound method, demonstrating that targeting the exact spectral norm yields an improved generalization performance compared to working with an upper bound.



\subsection{Robustness}

While generalization on a validation or test set gives an indication of model performance in practice,
data encountered in reality is often not as
exemplary as a curated benchmark data set and ensuring robustness against both natural and adversarial noise can often be a precondition for model deployment. 

As previously mentioned, earlier research has indicated that controlling the norm of the Jacobian is beneficial for robustness of our networks and we thus follow the path of \cite{DBLP:journals/corr/abs-1908-02729} and investigate how the robustness of the different schemes compare.
\begin{figure*}[h]
    \centering
    \includegraphics[width=\textwidth]{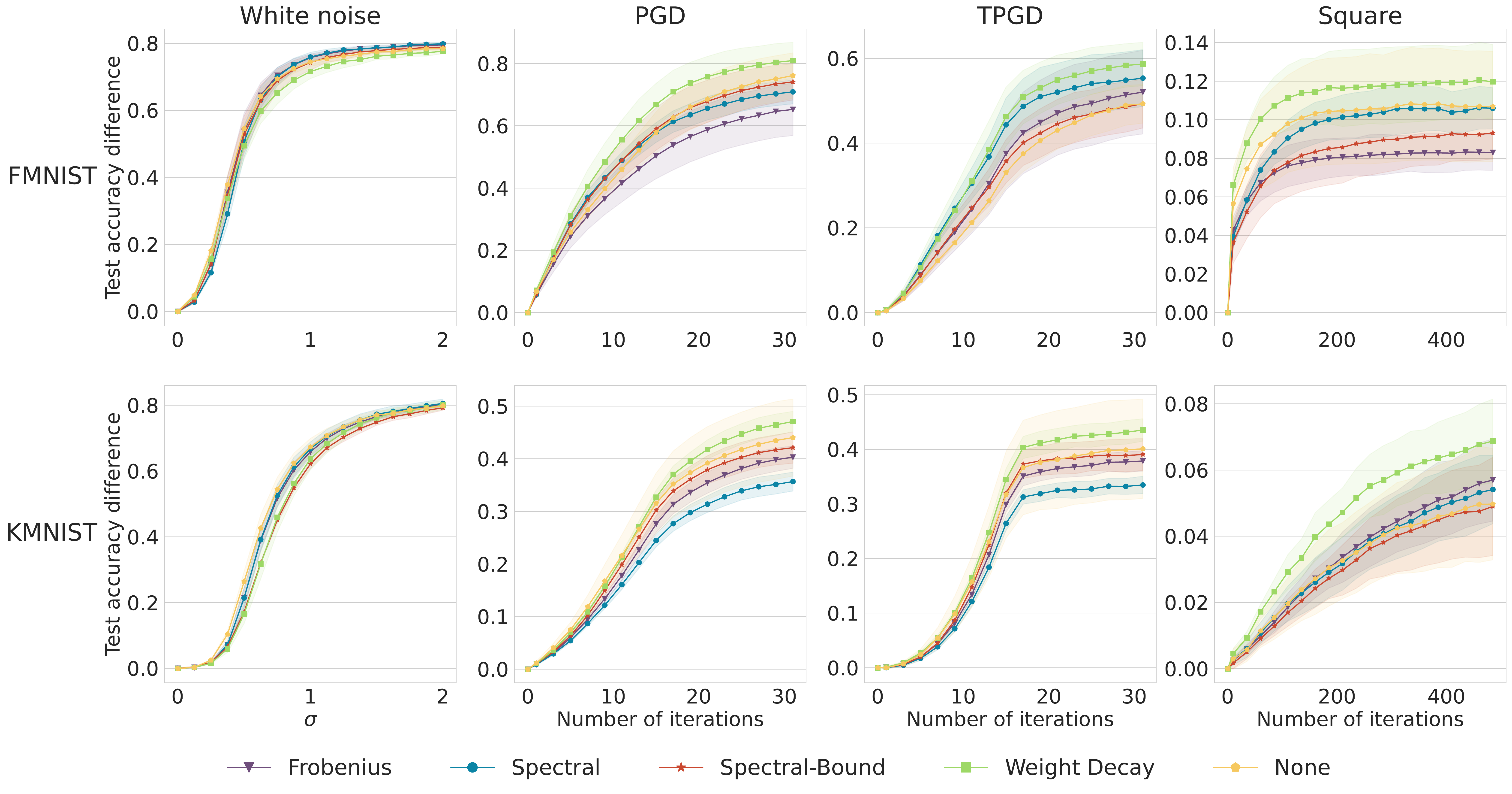}
    \caption{Robustness against perturbations. Each plot displays how the test accuracy drops as the perturbations gets stronger. Each column corresponds to a perturbation method and each row is associated with a given data set. The curves and intervals are obtained as the mean and standard deviation over 5 different networks.}
    \label{fig:adv_atk}
\end{figure*}

\subsubsection{Robustness against white noise}

We measure the robustness against white noise by creating a noisy validation set $\mathcal{D}_{v, \sigma^2}$ for FashionMNIST and KMNIST, consisting of data points $\tilde{x}$ obtained by adding independent Gaussian distributed noise to each individual pixel of validation points $x \in \mathcal{D}_v$ as
\begin{align}
    \tilde{x}_{ij} = x_{ij} + \epsilon,~~\epsilon \sim \mathcal{N}(0,\sigma^2)
\end{align}
whereupon we clip the value of all pixels into the range [0,1] and perform the aforementioned pre-processing.
Further, to enable a fair comparison between the different methods and to not have the result obscured by the initial baseline accuracies, we measure the difference between the baseline accuracy on $\mathcal{D}_v$ and the accuracy on $\mathcal{D}_{v,\sigma^2}$. These results can be seen to the left in Figure \ref{fig:adv_atk} where we see that there is not a large difference in robustness between either of the training schemes.

\subsubsection{Robustness against adversarial noise}

The last decade has seen an increased growth in the amount of research into adversarial noise, noise that may be imperceptible to the human eye but which has a considerable impact on the prediction of a deep learning model. Here we will work with the adversarial noise technique known as projected gradient descent (PGD) method \cite{DBLP:conf/iclr/MadryMSTV18} and related variants. PGD obtains the perturbation $\tilde{x}$ through a constrained gradient ascent, moving in a direction which increases the loss $l_{bare}(\theta, \{(x,y)\})$ while simultaneously restricting the ascent to the ball $B_{\delta} = \{z \in \mathbb{R}^{n_0}: ||x - z||_{\infty} \leq \delta\}$ which ensures that the perturbation $\tilde{x}$ is visually similar to $x$. Formally, PGD obtains $\tilde{x}$ as
\begin{align}
    \tilde{x} = \textrm{Proj}_{B_{\delta}}\bigg[x + \eta \textrm{sign}\bigg(\frac{d l_{bare}\big(\theta, \{(x,y)\}\big)}{dx} \bigg)\bigg],
    \label{eq:adv}
\end{align}
where Proj denotes the projection operator and $\eta$ the step-size for the gradient ascent. The gradient ascent process can be repeated over several iterations to yield perturbations $\tilde{x}$ indistinguishable from $x$ but for which the network predicts an incorrect label. After the ascent procedure we clip the pixel values into the range [0,1] and perform the pre-processing as before. We will additionally consider the adversarial attack methods
TPGD \cite{DBLP:conf/icml/ZhangYJXGJ19} that performs PGD on a Kullback-Leibler divergence of the softmax-scores, and the gradient-free attack Square \cite{DBLP:conf/eccv/AndriushchenkoC20}. All attacks are implemented through the torchattacks library \cite{kim2020torchattacks} with default parameters except for the parameters $\delta, \eta$ which we set to be 32/255 and 2/255 respectively.

To control the strength of the adversarial noise we vary the number of iterations for each attack. As before, to enable a fair comparison we compute the difference between the baseline accuracy on FashionMNIST and KMNIST with the adversarially perturbed validation sets. These results can be seen in Figure \ref{fig:adv_atk}
where we see that the Spectral regularization scheme is able to consistently ensure stronger robustness for the PGD and TPGD attacks on KMNIST compared to other methods, but that the full results indicate that no regularization algorithm is clearly dominant in all settings. Each regularization method (except weight decay) has some data set and attack where it outperforms the other methods in terms of mean test accuracy difference. Some form of Jacobian regularization is thus beneficial, but the exact penalization method to achieve an optimal robustness against adversarial noise is likely situation dependant.

\begin{figure}[h]
\centering

\begin{tikzpicture} [transform shape, scale = 1]

\tikzstyle{arrow} = [thick,->,>=stealth]

\newcommand\imgWidth{0.2}

\node[inner sep=0pt] (expandedImg) at (0, 0)
    {\includegraphics[width=0.45\textwidth]{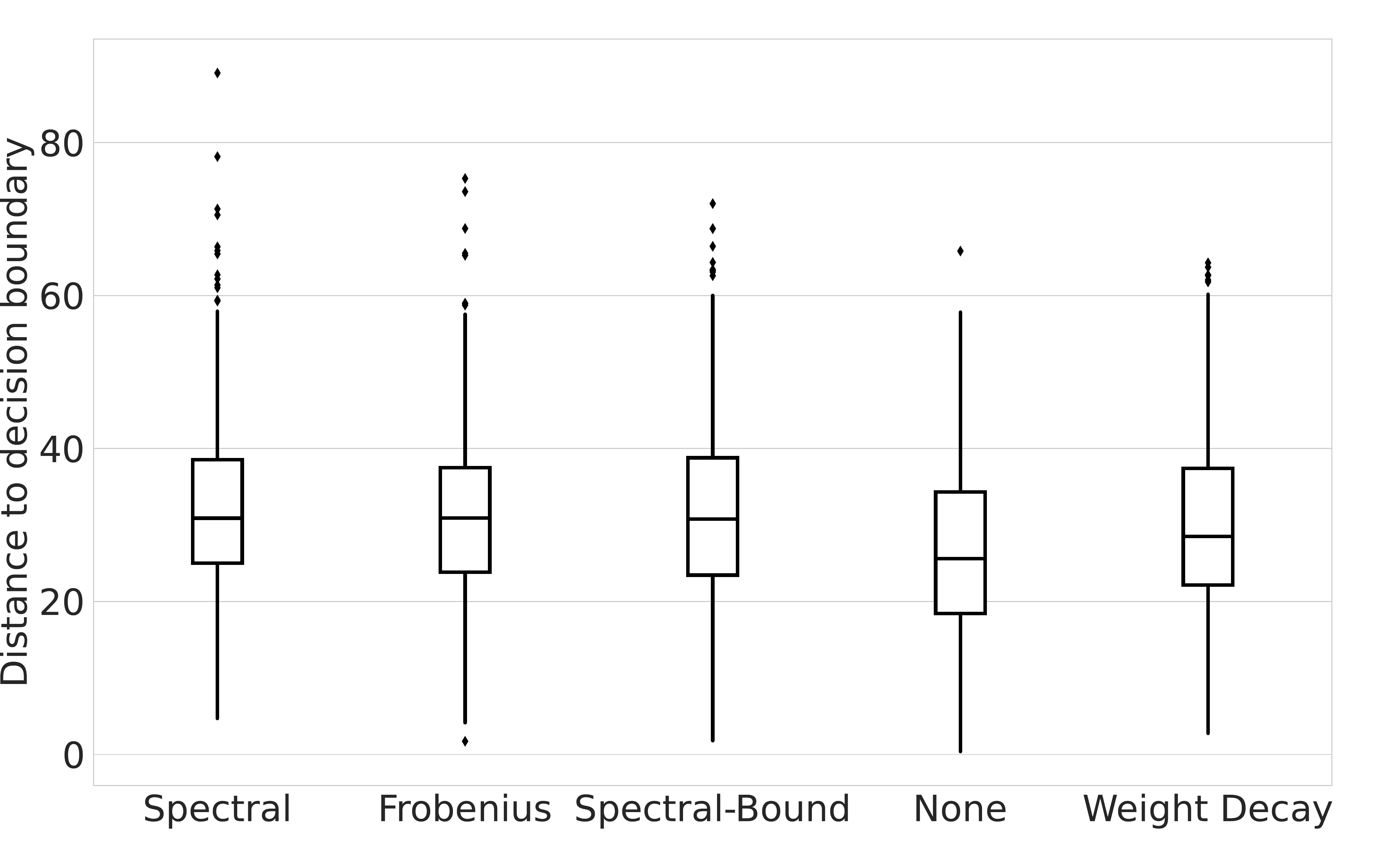}};
\node[inner sep=0pt] (expandedImg) at (0, -5)
    {\includegraphics[width=0.45\textwidth]{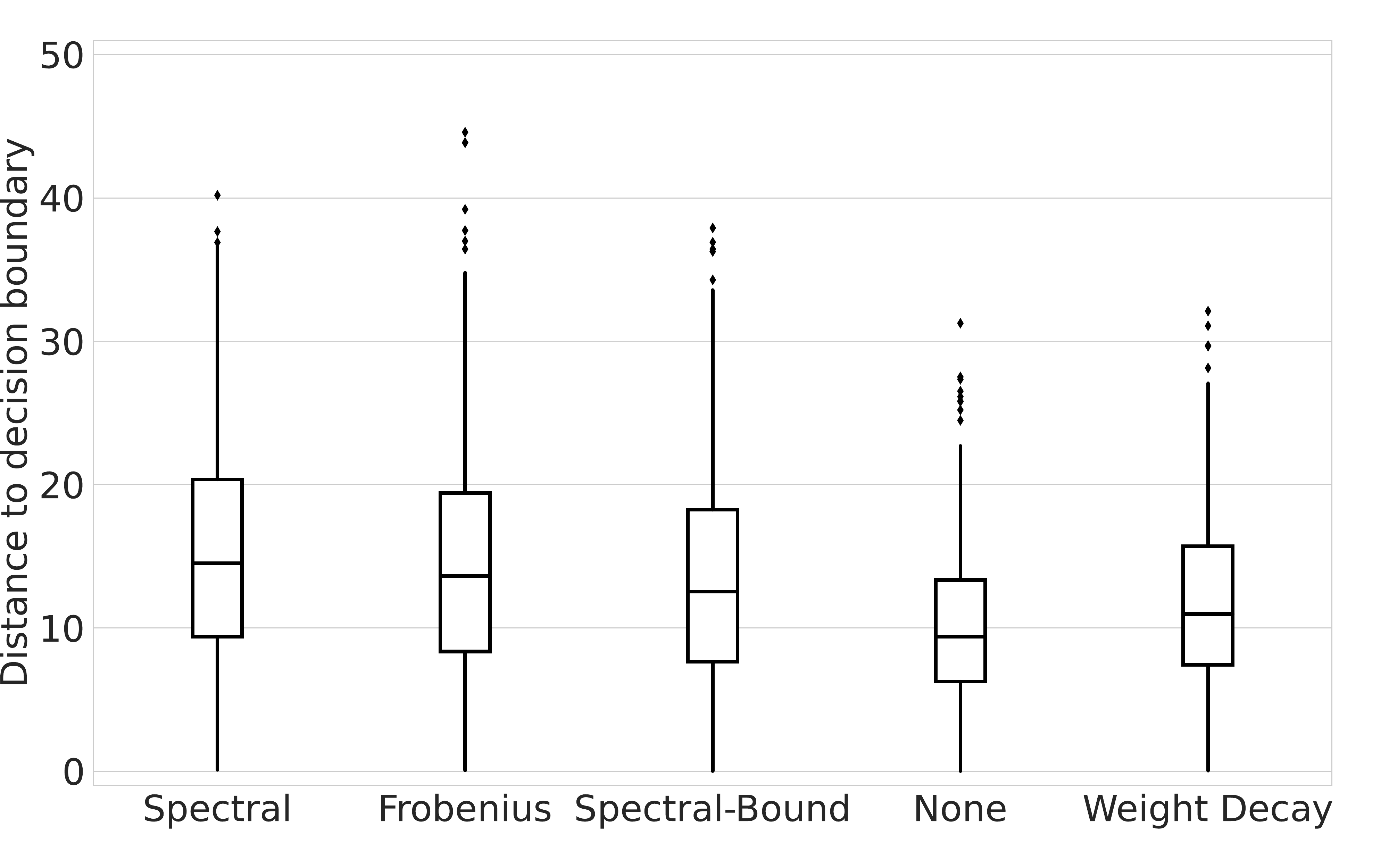}};

\end{tikzpicture}

\caption{Distance to the decision boundary. (Top) Distance to the nearest boundary for validation points in KMNIST. (Bottom) Distance to the nearest boundary for validation points in FashionMNIST. It can be seen that Spectral and Frobenius regularization increases the size of the decision regions on average.}
\label{fig:dist_b}
\end{figure}

\subsubsection{Distance to decision boundary}

One way to attempt to understand the robustness results is to analyze the distance to the closest decision boundary. Previous research demonstrated that controlling the Frobenius norm enlarged the decision cells and thus argued that this made the network more robust to perturbations \cite{DBLP:journals/corr/abs-1908-02729}.
We extend their experiments and perform an extensive investigation to measure the robustness where we measure the distance to the decision boundary for all validation points in FashionMNIST and KMNIST. To measure the distance to the decision boundary for a given point we sample points uniformly on concentric spheres of different radii and perform a binary search to find the smallest radii such that a sampled point obtains a predicted class different from the validation point at the center of the sphere. These results are summarized in Figure \ref{fig:dist_b}.

From these results we see that penalizing the spectral and Frobenius norm of the Jacobian yields regions which are larger than the other methods on average for FashionMNIST while all regularization methods achieve similarly sized regions for KMNIST. That some methods obtain similarly sized regions yet provide a varying level of safeguard against adversarial noise as seen in Figure \ref{fig:adv_atk} implies that the enlargening of the regions cannot fully capture the nuances of robustness against adversarial attacks. Choosing a model based on this intuition that larger regions provide a stronger safeguard can even yield a subpar model as evident from weight decays poor adversarial safeguard on KMNIST despite its large regions. 

We hypothesize that one aspect of robustness that this intuition fails to take into account is the structure of the loss landscape. Since a smooth loss landscape with large gradients will facilitate the creation of adversarial examples through gradient ascent we must also consider this aspect to get a holistic view of a regularization methods robustness.

\subsection{Time efficiency and relative error}
In the previous section we have detailed the generalization and robustness results when using the different regularization schemes. In this section we return to our initial task of investigating the difference between targeting the exact 
spectral norm of the Jacobian compared to working with an upper bound. From Table \ref{table:gen_res} we saw that this yields an improved generalization performance and from Figure \ref{fig:adv_atk} we observed that the two methods provide a similar protection against noise, with different strengths against different attacks on the two considered data sets.

While an improved generalization performance is beneficial, it cannot come at a too large of a computational cost. Additionally, with approximate methods it is also important to measure the trade-off between computational speed and accuracy of the approximated quantity. We thus analyze the computational overhead that they add to the training routine and the relative error with the analytical spectral norm.

\begin{figure}[h]
\centering

\begin{tikzpicture} [transform shape, scale = 1]

\tikzstyle{arrow} = [thick,->,>=stealth]

\newcommand\imgWidth{0.2}

\node[inner sep=0pt] (expandedImg) at (0, 0)
    {\includegraphics[width=0.49\textwidth]{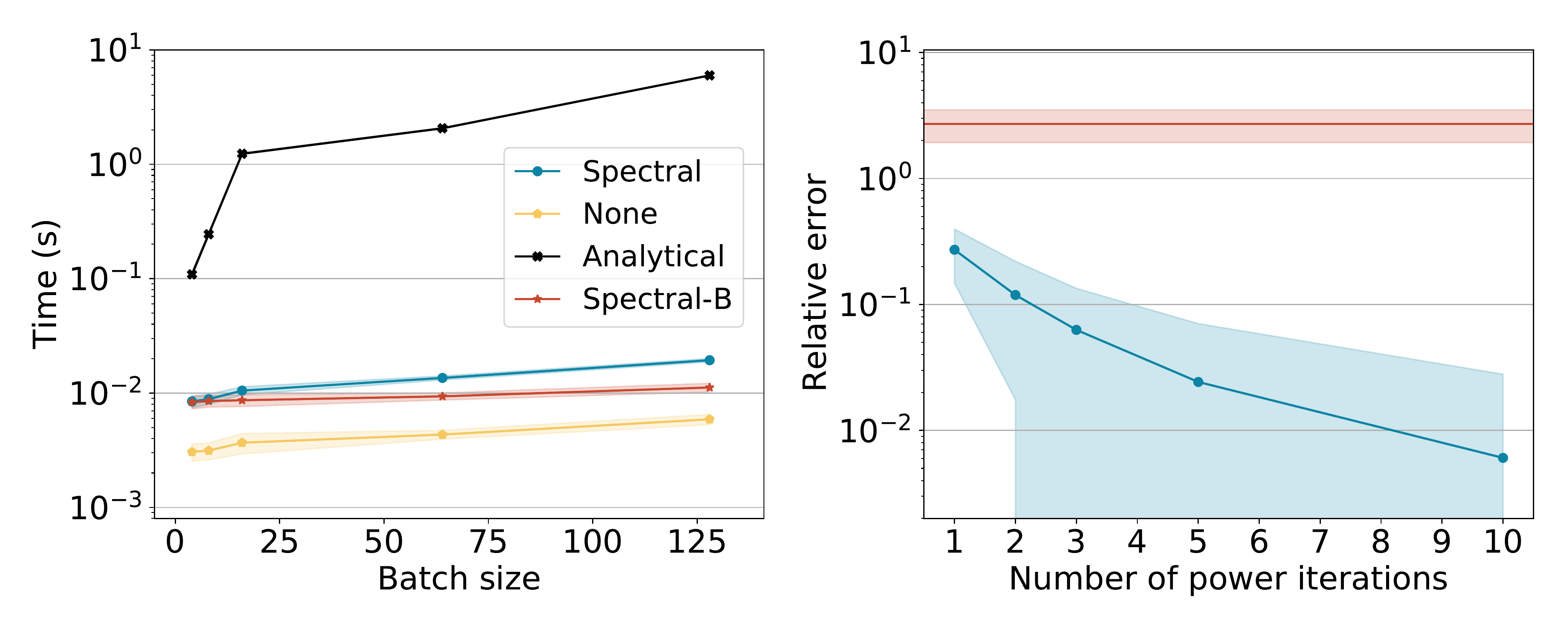}};

\node[scale = 0.7] at (3.4, 0.65) {$\prod_l ||W^l||_2$};

\draw [->] (2, 0.9) to [out=-30,in=180] (2.8, 0.65);

\end{tikzpicture}

\caption{Time and error comparison between the Spectral and Spectral-Bound method for the LeNet network. (Left) Time taken to pass over one batch of data points. The Spectral method is slower than the Spectral-Bound method for larger batch sizes but still around two orders of magnitude faster than calculating the exact spectral norm analytically. (Right) The relative error as the number of power iterations is increased. The relative error decreases quickly and is significantly closer to the exact quantity compared to the upper bound $\prod_l ||W^l||_2$.}
\label{fig:errtime}
\end{figure}

In Figure \ref{fig:errtime} (left) we can thus see the average time taken to optimize over a batch for the Spectral method, the Spectral-Bound method, an analytical method that calculates $||W_R||_2$ exactly and a regular forward-pass. In Figure \ref{fig:errtime} (right) the relative error for the power iteration scheme is visible.


From these plots we can see there is a small extra incurred cost of working with our method compared to regularizing with Spectral-Bound, but that our method has a significantly lower relative error while still being orders of magnitude faster than calculating the analytical spectral norm.

\section{Conclusion}

We have demonstrated a method to improve spectral norm regularization for neural networks. While previous methods relied on inexact upper bounding techniques, our technique targets the exact spectral norm. In the piecewise linear setting our method is easily implemented by performing power iteration through a forward-backward scheme while generally it can be achieved with a slightly more intricate scheme where the underlying computational graphs are modified to perform the power iteration scheme. 

This scheme obtained an improved generalization performance while achieving a similar safeguard to natural and adversarial noise as other Jacobian regularization techniques. Further, we investigated the intuition that Jacobian regularization provides a strong defence against adversarial attacks by the enlargening of the decision cells and found that the size of the regions is not necessarily indicative of the robustness of the network.

For future work we are interested in applying our scheme to more complex data sets and tasks, for example Reinforcement learning and Generative Adversarial Networks where controlling the spectral norm has already proven to be beneficial \cite{DBLP:conf/icml/GogianuBRCBP21, DBLP:conf/iclr/MiyatoKKY18}. We believe that our scheme can yield additional benefits and will spur further research into accurate and well-principled spectral and Jacobian regularization techniques.

\bibliography{biblio}
\bibliographystyle{icml2021}

\newpage
\appendix
\onecolumn
\section{Experimental details}
\subsection{Network architectures}
We will follow \cite{DBLP:journals/corr/abs-1908-02729} and denote a convolutional-max-pool layer as a tuple ($K$, $C_{in}\rightarrow C_{out}, S, P, M$) where $K$ is the width of the kernel, $C_{in}$ is the number of in-channels, $C_{out}$ the number of out-channels, $S$ the stride, $P$ the padding of the layer and $M$ the size of the kernel of the max-pool following the convolutional layer. The case $M=1$ can be seen as a convolutional layer followed by an identity function. Linear layers we will denote as the tuple ($N_{in}$, $N_{out}$) where $N_{in}$ is the dimension of the input and $N_{out}$ the size of the output. For KMNIST and FashionMNIST we used the LeNet network which consist of a convolutional-maxpool layer (5, $1 \rightarrow 6$, 1, 2, 2), convolutional-maxpool layer (5, $6 \rightarrow 16$, 1, 0, 2), linear layer (400, 120), linear layer (120 , 84) and linear layer (84, 10).

We use the VGG16 network as is available from the torchvision package. For this network we use batch-norm layers directly after every convolutional layers. This network consist of the layers  (3, $3 \rightarrow 64$, 1, 1, 1), (3, $64 \rightarrow 64$, 1, 1, 2), (3, $64 \rightarrow 128$, 1, 1, 1), (3, $128 \rightarrow 128$, 1, 1, 1), (3, $128 \rightarrow 256$, 1, 1, 2), (3, $256 \rightarrow 256$, 1, 1, 1), (3, $256 \rightarrow 256$, 1, 1, 1), (3, $256 \rightarrow 512$, 1, 1, 2), (3, $512 \rightarrow 512$, 1, 1, 1), (3, $512 \rightarrow 512$, 1, 1, 1), (3, $512 \rightarrow 512$, 1, 1, 2), (512, 10).

\subsection{Training details and code}
We train the LeNet networks for 50 epochs with SGD (with momentum=0.8). For every regularization method we perform a hyperparameter search over these three following parameters and values.
\begin{itemize}
    \item Learning rate: [0.01, 0.001]
    \item Batch size: [16, 32]
    \item Weight factor $\lambda$: [0.0001, 0.001, 0.01, 0.1]
\end{itemize}
For the VGG16 network we trained the network for 100 epochs with a batch size of 128, SGD with momentum of 0.8 and performed a hyperparameter search over these parameters and values
\begin{itemize}
    \item Weight factor $\lambda$: [0.00001, 0.0001, 0.001, 0.01, 0.1]
\end{itemize}
For VGG16 we additionally used a cosine annealing learning rate scheduler with an initial learning rate of 0.1 and the data augmentation techniques of random cropping and horizontal flipping.

For each hyperparameter setting we repeat the training procedure 5 times to be able to obtain mean and standard deviation. We pick the final representative model for each regularization method as the one that achieves the lowest mean validation loss over these 5 training runs.

For the Frobenius regularization we set $n_{proj} = 1$ and for the Spectral-Bound we estimate the spectral norm of the weight matrices through one power iteration.


\subsection{Details for figures}

\textbf{Figure 4:} The model for each regularization method was chosen randomly among the 5 models from the hyperparameter setting that obtained the best results in Table 1. The distance is only calculated for the points in the validation set that all models predict correctly. In total the distance is predicted for between 8000 - 9000 validation points on FashionMNIST and KMNIST.

\textbf{Figure 5 (left):} The time for a batch was measured on a computer with NVIDIA K80 GPU as available through Google Colab\footnote{colab.research.google.com}. The analytical method works by sequentially calculating $d(x^L \cdot e_i)/dx$ where $e_i$ is a basis-vector for $\mathbb{R}^{n_{out}}$ for $i=1,2,...,n_{out}$. This yields the full Jacobian matrix which we then calculate the singular values of by using inbuilt functions in PyTorch.

\textbf{Figure 5 (right):}~~~~The upper bound was evaluated on a network trained with the Spectral-Bound regularization scheme for all data points in the training set. The curve for the spectral method was evaluated on a network trained with the spectral method for all data points in the training set. For the spectral method there was no significant difference in the shape of the curve when using a different network or by working with data points in the validation set.

\section{Conversion between operators}
In this section we detail how to convert between the forward $F$, backward $F^T$ and regular operators $G$. These can be seen in Table \ref{tab:linear} - \ref{tab:max_pool}. Other non-linearities such as Dropout can be incorporated identically to ReLU by simply storing the active neurons in a boolean matrix $Z$. 


\subsection{Skip-connections}
Utilizing networks with skip-connections does not change the forward and backward modes. Simple turn off the bias of all layer transformations and replace the activation functions with the matrices $Z_R^i$ instead. That this is true follows from the definition of a network with skip-connections. For simplicity of presentation, we will assume that the skip-connections only skip one layer. Assume that we have a network with $L$ layers and additionally have skip-connections between layers with indices in the set $\mathcal{S} := \{s_1,s_2,...,s_m\},~1\leq s_i \leq L$. Then the network $f_{\theta}$ is given recursively as before with
$$
x^l = \begin{cases} f^l(G^l(x^{l-1}) + b^l) &\mbox{if } l \in \mathcal{S}^C, \\
x^{l-1} + f^l(G^l(x^{l-1}) + b^l) & \mbox{if } l \in \mathcal{S}. \end{cases}
$$
Assuming that we are only working with piecewise linear or linear operators $G^l$, then for $x \in R$
we know that each operator can be represented as a matrix and we can write the derivative of the two cases as
$$
\frac{dx^l}{dx^{l-1}} = \begin{cases} Z^lW^l &\mbox{if } l \in \mathcal{S}^C, \\
I + Z^lW^l & \mbox{if } l \in \mathcal{S}, \end{cases}
$$
where $I$ denotes a unit-matrix. The Jacobian-vector product $W_Rv$ can thus be obtained as
\begin{align}
 W_Rv =  \bigg(\prod_{l=1}^L (I - \mathbb{I}\{l \in \mathcal{S}^C\} + Z^lW^l)\bigg)v
\end{align}
where $\mathbb{I}\{l \in \mathcal{S}^C\}$ is an indicator for the unit-matrix so that we can concisely write the two cases. Thus we see that we can interpret this equation in the same manner as we did for the networks without skip-connections. We simply pass the input $v$ through the network and turn off all the biases and replace the activation functions with $Z^l$. The same is true for the backward mode.
\begin{center}
\begin{table*}[h]
    \centering
    
\begin{tabular}{ p{3.5cm}|p{6cm}|p{3.5cm} } \hline
 Forward-pass ($G$) & Forward-mode ($F$) & Backward-mode ($F^T$) \\ 
 \hline
 \textbf{Input:} $x,W,b$\newline
  $y$ = Linear($x,W,b$) \newline
  \textbf{return:} $y$ & 
  \textbf{Input:} $x,W$\newline
  $y$ = Linear($x,W,0$)\newline
     \textbf{return:} $y$
    & \textbf{Input:} $x,W$\newline
    y = Linear($x,W^T,0$)\newline
    \textbf{return:} y\\
    \hline
\end{tabular}
\caption{Conversion table for the linear operator.}
    \label{tab:linear}
\end{table*}
\end{center}

\begin{center}
\begin{table*}[h]
    \centering
\begin{tabular}{ p{3.5cm}|p{6cm}|p{3.5cm} } \hline
 Forward-pass ($G$) & Forward-mode ($F$) & Backward-mode ($F^T$) \\ 
 \hline
 \textbf{Input:} $x,W,b$\newline
  $y$ = Conv($x,W,b$) \newline
  \textbf{return:} $y$ & 
  \textbf{Input:} $x,W$\newline
  $y$ = Conv($x,W,0$)\newline
     \textbf{return:} $y$
    & \textbf{Input:} $x,W$\newline
    y = ConvTranspose( $x,W,0$)\newline
    \textbf{return:} y\\
    \hline
\end{tabular}
\caption{Conversion table for the convolutional operator.}
    \label{tab:conv}
\end{table*}
\end{center}

\begin{center}
\begin{table*}[h]
    \centering
\begin{tabular}{ p{3.5cm}|p{6cm}|p{3.5cm} } \hline
 Forward-pass ($G$) & Forward-mode ($F$) & Backward-mode ($F^T$) \\ 
 \hline
 \textbf{Input:} $x$\newline
  $y$, indices = maxpool($x$) \newline
  $I$ = indices\newline
  $S$ = x.shape\newline
  \textbf{return:} $y,I,S$ & 
  \textbf{Input:} $x$, $I$\newline
  y = x[I]\newline
  \textbf{return:} $y$
    & \textbf{Input:} $x,I,S$\newline
    y = maxunpool(x,  indices=$I$,   shape = $S$)\newline
    \textbf{return:} y\\
    \hline
\end{tabular}
\caption{Conversion table for the max-pool operator.}
    \label{tab:max_pool}
\end{table*}
\end{center}

\section{Proof for extension scheme}
We will denote the directed acyclic graph which when summing the product of every edge element along every path from output to input yields $(df/dx)^T$ as $G$.

\textbf{Theorem:} \textit{Consider the graph F obtained by flipping the direction of all edges of G and adding a node at the end of F with edge elements given by components of v.
Summing the product of every edge element along every path from output to input of $F$ yields $(df/dx)v$.}

\textbf{Proof:} We will follow the notation of \cite{lecnotesjac}, Theorem 1 and denote the Jacobian between variables $y=f_{\theta}(x)$ and $x$ as the sum of the product of all intermediate Jacobians, meaning
\begin{align}
    \frac{dy}{dx} = \sum_{p \in \mathcal{P}(x,y)} \prod_{(a,b) \in p} J^{a\rightarrow b}(\alpha^b)
\end{align}
where $\mathcal{P}(x,y)$ is the set of all directed paths between $x$ and $y$ and $(a,b)$ is two successive edges on a given path.

In our scheme we flip the direction of all relevant edges and add a fictitious node at the end of the path the flipped paths. Since we preserve the edge elements, we can realize that flipping the direction of the edges simply transposes the local Jacobian, meaning that $J^{b\rightarrow a}(\alpha^b) = \big(J^{a\rightarrow b}(\alpha^b)\big)^T$ with our scheme. Further, our added fictitious node has edge elements given by elements of $v$, and the Jacobian between that node and the subsequent layer is thus given by $v^T$.
For a path $p=[(v_1,v_2), (v_2,v_3),...,(v_{n-1}, v_n)]$ we define the flipped path with the added fictitious node as $p^T$ as $p^T = [(v_n, v_{n-1}), ..., (v_2,v_1), (v_1, v_f)]$ and the reverse-order path $\neg p$ as $\neg p = [( v_{n-1}, v_n), ..., (v_2,v_3),(v_1,v_2)]$.
For our modified graph we thus have the Jacobian for a path as

\begin{align}
    \prod_{(a,b) \in p^T} J^{a\rightarrow b}(\alpha^b) &= 
    \prod_{(a,b) \in p^T} \big(J^{b\rightarrow a}(\alpha^b)\big)^T \\
    &=v^T\bigg(\prod_{(a,b) \in p} J^{a\rightarrow b}(\alpha^b)\bigg)^T\\
    &=v^T\bigg(\prod_{(a,b) \in \neg p} J^{a\rightarrow b}(\alpha^b)^T\bigg)
\end{align}

Denoting the fictitious node as $n_f$ and summing over all paths we thus get
\begin{align}
    &\sum_{p^T \in \mathcal{P}(y,n_f)} \prod_{(a,b) \in p^t} J^{a\rightarrow b}(\alpha^b) \\
    &= \sum_{p^T \in \mathcal{P}(y,n_f)}v^T\bigg(\prod_{(a,b) \in \neg p} J^{a\rightarrow b}(\alpha^b)^T\bigg)\\
    &=v^T\sum_{p^T \in \mathcal{P}(y,n_f)}\bigg(\prod_{(a,b) \in \neg p} J^{a\rightarrow b}(\alpha^b)^T\bigg)\\
    &=v^T\big(\frac{dy}{dx}\big)^T = (\frac{dy}{dx}v)^T
\end{align}
which proves that working with the modified graph will yield the desired matrix-vector product $\frac{dy}{dx}v$ $\square$.

\end{document}